\documentclass[sigconf]{acmart}
\AtBeginDocument{%
  \providecommand\BibTeX{{%
    \normalfont B\kern-0.5em{\scshape i\kern-0.25em b}\kern-0.8em\TeX}}}

\setcopyright{acmcopyright}
\copyrightyear{2018}
\acmYear{2018}
\acmDOI{10.1145/1122445.1122456}

\begin{document}

\title{A machine learning pipeline for aiding school identification from child trafficking images}

\author{Sumit Mukherjee \\
Tina Sederholm \\
Anthony C. Roman \\
Ria Sankar}
\email{{summukhe,tinase,ancintro,rias}
@microsoft.com}
\affiliation{%
  \institution{Microsoft Corporation}
  \city{Redmond}
  \country{USA}
}

\author{Sherrie Caltagirone}
\email{sherrie@globalemancipation.ngo}
\affiliation{%
  \institution{Global Emancipation Network}
  \country{USA}
}

\author{Juan Lavista Ferres}
\email{jlavista@microsoft.com}
\affiliation{%
  \institution{Microsoft Corporation}
  \city{Redmond}
  \country{USA}
}

\renewcommand{\shortauthors}{Mukherjee et al.}

\begin{abstract}
Child trafficking in a serious problem around the world. Every year there are more than 4 million victims of child trafficking around the world, many of them for the purposes of child sexual exploitation. In collaboration with UK Police and a non-profit focused on child abuse prevention, Global Emancipation Network, we developed a proof-of-concept machine learning pipeline to aid the identification of children from intercepted images. In this work, we focus on images that contain children wearing school uniforms to identify the school of origin. In the absence of a machine learning pipeline, this hugely time consuming and labor intensive task is manually conducted by law enforcement personnel. Thus, by  automating aspects of the school identification process, we hope to significantly impact the speed of this portion of child identification. Our proposed pipeline consists of two machine learning models: i) to identify whether an image of a child contains a school uniform in it, and ii) identification of attributes of different school uniform items (such as color/texture of shirts, sweaters, blazers etc.). We describe the data collection, labeling, model development and validation process, along with strategies for efficient searching of schools using the model predictions. 
\end{abstract}




\maketitle

\section{Introduction}

According to Human Rights First~\cite{hr}, the total number of victims of global human trafficking is approximately 24.9 million, with 5.5 million (25 percent) being children. Additionally, the percentage of children impacted has nearly doubled in the last 15 years (based on the 2020 UNODC Global Report on Trafficking in Persons~\cite{bouche2020unodc}). Operations of this criminal industry are not easily identifiable, and organizations like the Global Emancipation Network (GEN) identify patterns and build tools to aid law enforcement with taking down human trafficking operations globally. In this project we partnered with GEN to provide tools that would enhance the ability of authorities to take meaningful action quickly.

It is pointed out by our law enforcement partners that a large number of images scraped from child trafficking websites and from devices seized from child traffickers contain children in school uniforms. Since uniforms may provide information about the location where the child was abducted from, law enforcement officers currently try to manually identify the school based on descriptive characteristics of the school uniforms (such as color/texture of uniform elements). Speeding up this process is crucial as criminology experts believe that the first 72 hours is critical when a person has gone missing~\cite{abc}. In order expedite the process of school identification from images, we built a proof-of-concept machine learning pipeline to: i) identify whether an image contains a child wearing a school uniform (uniform prediction), and ii) to identify which school the uniform represents using attribution specific to that school (such as color of shirts, sweaters, blazers etc.). . 

In this paper we describe the current training data acquisition, labeling, model development, model validation and deployment approaches. We also outline some of the current shortcomings of our proof-of-concept pipeline as well as ideas for different school search strategies that can use our models outputs. Moving forward, in conjunction with efforts focused on gathering images from a large cross section of schools, this tool could greatly help law enforcement in the task of school identification in images of children. 

\section{Related work}
Identifying clothing items and their attributes from images is an important problem in computer vision. The earliest applications of this was in the e-commerce industry, with notable examples such as visual search tools of Pinterest~\cite{jing2015visual,zhai2019learning} and Amazon~\cite{amazon}. However, such tools were generally trained on massive proprietary datasets and are aimed at identifying semantically similar products/items which is a somewhat different goal than us.

To overcome the challenge of lack of public datasets, the last few years saw the publication of several large datasets of clothing items with rich categorical and segmentation labels~\cite{liu2016deepfashion,guo2019imaterialist}. ~\cite{liu2016deepfashion} is now widely used as a standard benchmark dataset for various computer vision tasks related to clothing classification, segmentation of clothing items, as well as generation of synthetic fashion images. While~\cite{guo2019imaterialist} is used in a highly successful Kaggle competion on image segmentation called iMaterialist~\cite{imat}.

While several models have been highly successful at the semantic segmentation of clothing items on these benchmark fashion datasets, there are several difficulties with directly applying such models to our setting. Firstly, the semantic labels (clothing item categories) of these benchmark datasets is usually much more fine grained than those required for uniform attribute prediction. Secondly, since the goal of our task is identification of attributes of clothing items and not the location of the items themselves, the existing models are not directly usable in our setting. Although, an accurate semantic segmentation model could be used as an object detection tool, which could then simplify the task of object attribute detection. Finally, due the staged nature of the benchmark images and the drastic difference in image size/quality between these and school uniform images available to us, we found that the pre-trained semantic segmentation models on these benchmark datasets to work rather poorly on our images. Hence, we focused on acquiring our own labeled dataset, and developed machine learning models specific to our task.  

\section{Machine learning pipeline overview}
Our machine learning pipeline is modeled after the decision making process of human law enforcement officers. In the absence of automation tools, officers first screen images for presence of uniforms and then identify uniform attributes, which are then used to look up schools. Geographical information about location of acquisition of images is also utilized to filter schools, since a large number of schools may have similar looking uniforms. In this initial pipeline, we only focus on the first two tasks undertaken by the law enforcement officers. To replicate the process undertaken by the officers, the goal of our modeling endeavor was several fold: i) to identify whether an image contains a child wearing a school uniform, ii) to identify which school the uniform belongs to, and iii) to do so in a manner that can scale to new schools. To achieve this we built a machine learning pipeline comprising of two modules: i) a uniform presence detection (uniform prediction) module, and ii) a uniform attribute identification module. Below, we describe each machine learning module.

\subsection{Uniform classification model}
\begin{figure}[h!]
\centering
    \includegraphics[scale=0.38]{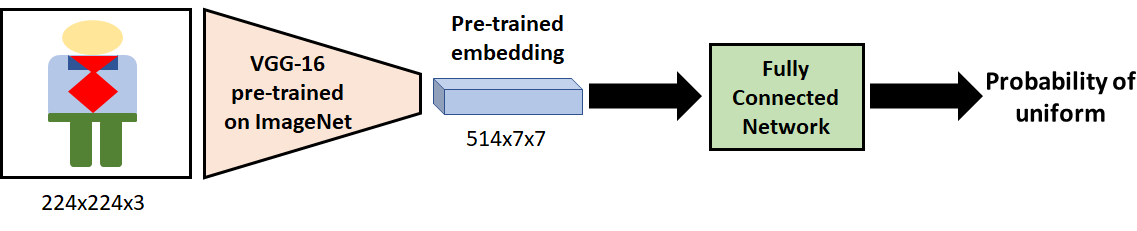}
    \caption{Overview of uniform classification module. The image is first passed through a pre-trained VGG16 model (trained on ImageNet). The output of the pre-trained model (from an embedding layer) is then flattened and passed through a fully connected network, which outputs a scalar probability as the output.}
    \label{fig:UniformClassifier}
\end{figure}
The uniform classification task is posed as a binary prediction task, where the model simply outputs whether it detects a uniform in the image (class 1) or does not detect a uniform in the image (class 0). To achieve this, we use a developed deep learning based model that takes a $224\times224$ pixel RGB image and produces the probability of an image belonging to class 1 as the output. We utilized the output of a  pre-trained machine learning model called VGG16~\cite{qassim2018compressed} which was trained on a large image classification dataset called ImageNet~\cite{deng2009imagenet}. This output was then passed through as the input to a fully connected network which output the probability of an image containing an uniform. The use of a pre-trained model in conjunction with a trainable model (the fully connected network, in this case) is common practice in data sparse classification scenarios like ours. This has been shown to significantly improve the classification performance in many settings~\cite{huh2016makes}. Note that VGG16 one of many pre-trained models that can be used for this task. Other pre-trained models such as InceptionV3 or Resnet-50 can also be used here and were seen to have similar performance. Also note that the image size of $224\times224$ pixel was chosen for this stage because that is the expected input image size for VGG16. 

\subsection{Uniform attribute prediction model}
\begin{figure}[!htb]
\centering
    \includegraphics[scale=0.32]{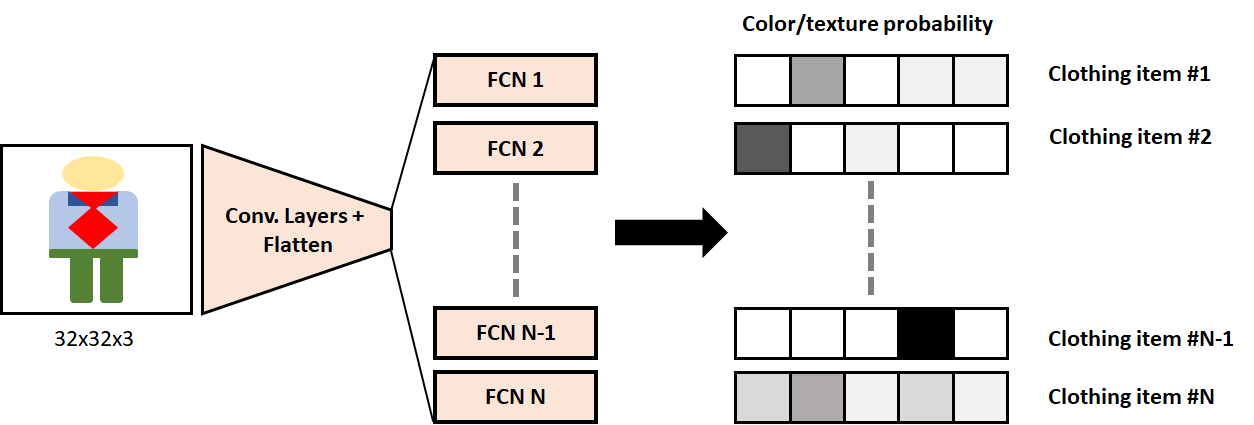}
    \caption{Overview of multi-label multi-class attribute prediction module. The image is first passed through a series of convolutional layers and then flattened. The flattened vector is them passed through several fully connected layers (abbreviated as FCN here), one for each clothing item. The output of the fully connected layers is a vector of probabilities of each color.}
    \label{fig:AttributeClassifier}
\end{figure}
The goal of the uniform attribute prediction model is to predict characteristics of certain clothing items that describe school uniforms. Due to limited training data, we have limited out clothing item categories to: i) Shirt, ii) Trousers, iii) Outer coat, iv) Jumper/Sweater/Cardigan, v) Dress, vi) Tie. The list of clothing items is expected to expand slightly, once we have more training data. In the current iteration of the model, we will be predicting the base colors for these clothing items from the following colors: i) Red/Brown, ii) Yellow/Orange, iii) Green, iv) Blue/Purple, v) White, and vi) Black/Grey, vii) No color (meaning that the clothing item is not present in the image). In future iterations of the model (once we have more training data) we will also including textures (such as stripes, polka dots etc.). The current model architecture is a multi-label multi-class deep learning model comprising of several convolutions layers followed by separate fully connected layer networks for each clothing item. The output of each FCN module, is the probability for each color (and absence of the clothing item). Note that in our multi-label multi-class setting, the clothing items are the labels and the colors are the classes.

\section{Data acquisition and labeling}
\subsection{Uniform classification}
The uniform classification model was trained on a total of 2000 images. Half of the images were images of children in casual wear collected by the GEN volunteers from their friends/family (with their permission). The remaining images were from 10 UK schools  of children wearing school uniforms (100 images each) with their permission. All images had the faces of the children removed. While the dataset was balanced in terms of labels, the dataset was somewhat imbalanced in other ways such as gender or uniform types. In this initial prototype, due to limited availability of labeled data, we ignore such considerations but as we have more data we can consider approaches such as stratified sampling to create a well balanced training dataset. 

\subsection{Uniform attribute prediction}
For the uniform attribute prediction task, several thousand images were collected by scraping websites of $80$ randomly chosen schools in the UK. Individual persons were then identified and separated in each image using a publicly available object recognition tool called Mask-RCNN ~\cite{matterport_maskrcnn_2017}. This step was done since most images of trafficked children contain only one child according to law enforcement. This led to approximately ten thousand images containing single individuals. Multiple volunteers from GEN (Global Emancipation Network) then used Azure Labeling Services~\cite{al} to label each image with the afore mentioned clothing and color combinations (texture was also collected but not used in the current modeling). To reduce human errors, each label was verified by at least two volunteers. Images that were of very poor resolution were discarded, which left four thousand usable images.

\section{Model evaluation}
\subsection{Uniform classification}
For the uniform classification model, we performed two separate validations. The first validation focused on the models performance on a held out test dataset. To this end, we randomly selected $80\%$ of the available data for training our deep learning model(s) and the remaining $20\%$ for model evaluation. This $20\%$ was not used in any way during the model training process. We found the uniform prediction model to have a $96\%$ accuracy on it's held-out test dataset. This is a reasonably high number for a balanced binary classification task, where the baseline accuracy is $50\%$. 

\begin{figure}[h!]
\centering
    \includegraphics[scale=0.6]{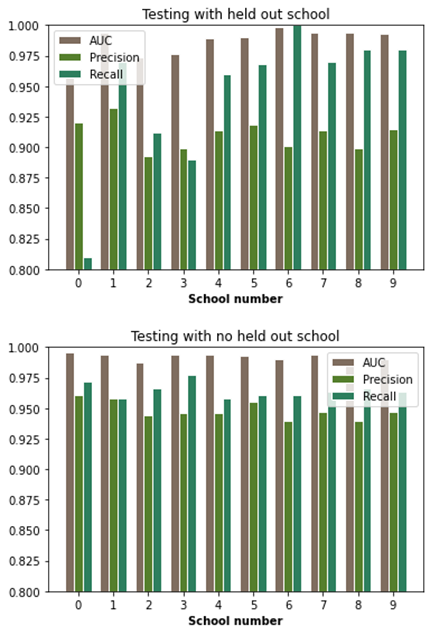}
    \caption{Comparison of uniform classification model performance using different metrics on two different test scenarios: (top) test sets are a randomly held out split of the data, (bottom) models are trained on all but one school and test set contains the one school that wasn't used during model training.}
    \label{fig:UC_Results}
\end{figure}

For the next validation, we wanted to test robustness of the uniform classification model on schools it has not scene before. To this end, we created 10 training sets leaving one school out in each i.e. using 9 schools. We then created 2 separate groups of tests sets: i) 10 test sets using non-uniform images and uniform images from the same 9 schools as each training set, ii) 10 test sets using non-uniform images and uniform images from the left out school from each training set. The ratio of uniform to non-uniform images was kept constant in both cases. We report the results in Figure~\ref{fig:UC_Results}, demonstrating that our uniform classification models perform similarly well on both test scenarios indicating robust performance on unseen schools. Despite the results showing model robustness, it should be noted that due to the limited number of schools used in training, such a model may not generalize to all unseen schools. However, we note here that the goal of this work is to develop a proof-of-concept prototype and in a real-world deployment scenario the model will be trained on a much broader set of schools.

\subsection{Uniform attribute prediction}
\begin{figure}[!htb]
\centering
    \includegraphics[scale=0.3]{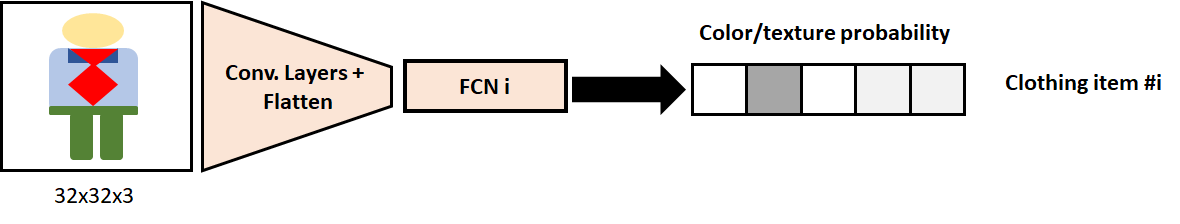}
    \caption{Overview of single-label multi-class attribute prediction module. The image is first passed through a series of convolutional layers and then flattened. The flattened vector is them passed through a fully connected layer module. The output of the fully connected layers is a vector of probabilities of each color for clothing item $i$. The convolutional and fully connected layer module architectures are identical to the multi-label multi-class model.}
    \label{fig:AttributeClassifier_single}
\end{figure}
For the attribute prediction task, we randomly selected $80\%$ of the available data for training our model and the remaining $20\%$ for model evaluation. Since 
each clothing category has multiple color options and the relative abundance of each color is different in the dataset, the baseline accuracy is no longer $50\%$. In such an imbalanced multi-label prediction problem, there are variety of alternative baselines possible. We picked the baseline that predicts colors randomly with the same proportion as the colors' abundance in the test dataset (for each clothing item independently). Another baseline was a single label multi-class model (Figure~\ref{fig:AttributeClassifier_single}) for each clothing item (label) seperately. The the convolutional and FCN modules were identical to the multi-label multi-class model. 
\begin{figure}[h!]
\centering
    \includegraphics[scale=0.35]{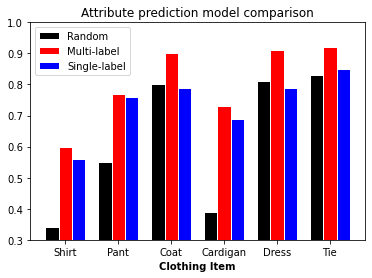}
    \caption{Comparison of attribute prediction model performance with different baselines.}
    \label{fig:AP_Results}
\end{figure}
As seen in Figure~\ref{fig:AP_Results}, our multi-label multi-class model does better than both baselines on all the clothing item categories. For several clothing items (such as shirts or sweaters) we see a significant improvement over the random baseline. Not surprisingly, clothing items where one color (or absence of the item) doesn't dominate the examples, our model shows a larger improvement over the random baseline. This indicates that as we will have more examples of each color for each clothing item, the model performance is likely to significantly improve over the baseline. To this end, Global Emancipation Network has collected approximately one hundred thousand more images (which are yet unlabeled). Adding these images to the training dataset is likely to significantly improve the model.

\section{Deployment and known limitations}
We deployed out model as a docker container that is capable of running entirely offline. The major known limitation of our current model is that the predictions are poor for large images (since most of the training images were relatively small ~20-100 kb). We hope to remedy this by re-training the model on a larger and more diverse set of images. 

\section{Future work}
We are currently working on developing a continuous update pipeline for our machine learning pipeline. Briefly, the goal is to enable continuous re-training of the model as more data and labels are collected. 

While this paper has focused on developing the machine learning portion of the proposed school identification framework, another related but important component is the school search using the predictions of the machine learning models. Notably, using approaches to make the search process fault tolerant, could substantially reduce the human verification required for our predicted results.

\section{Conclusion}
In this study, we demonstrate how the use of a well-defined machine learning pipeline can expedite the work of local law enforcement authorities by identifying a specific school uniform. Our proof-of-concept pipeline comprises of a uniform classification to identify whether an image contains a child in uniform, and a uniform attribute prediction model which predicts the color (or absence) of different uniform relevant clothing items. 

We explained the rationale for choosing such a pipeline as well as our choice of the various models. Furthermore, we tested the different components of our pipelines on various test scenarios and showed superior performance against relevant baselines. Finally, we explain the deployment process as well as known issues with our pipeline.

The methodology and classification models presented in this study can easily be expanded to regions where school uniforms are common, such as India. However, the adoption of such a pipeline might face challenges in countries where school uniforms are uncommon. We anticipate the current pipeline provides limited ability for law enforcement to re-use the model for the other purposes. However, due to valid concerns about re-usability of machine learning models by law enforcement for reasons other than ones they were developed for, any practical deployment of this pipeline will first go through legally binding 'limited use' agreements with agencies using the pipeline. 

\bibliographystyle{ACM-Reference-Format}
\bibliography{example_paper}

\end{document}